%% file: main.tex
\documentclass[11pt,a4paper]{article}
\usepackage[hyperref]{emnlp2018}
\usepackage{times}
\usepackage{latexsym}
\usepackage{amsfonts,amsthm,amsmath,amssymb}
\usepackage{url}
\usepackage{graphicx}
\usepackage{booktabs}
\usepackage{caption}
\aclfinalcopy % Uncomment this line for the final submission

\newcommand{\Z}{\mathbb{Z}}
\newcommand{\Q}{\mathbb{Q}}
\newcommand{\R}{\mathbb{R}}

\renewcommand{\epsilon}{\varepsilon}

\title{Neural Linguistic Steganography}

\author{Zachary M. Ziegler\thanks{$\;\;$Equal contribution.} $\;\;\;\;$ Yuntian Deng\footnotemark[1] $\;\;\;\;$ Alexander M. Rush \\
  School of Engineering and Applied Sciences \\ Harvard University, Cambridge, MA \\
  {\tt \{zziegler@g, dengyuntian@g, srush@seas\}.harvard.edu} }

\begin{document}
\input{preamble.tex}

\maketitle

\begin{abstract}
Whereas traditional cryptography encrypts a secret message into an unintelligible form, steganography conceals that communication is taking place by encoding a secret message into a cover signal. Language is a particularly pragmatic cover signal due to its benign occurrence and independence from any one medium. Traditionally, linguistic steganography systems encode secret messages in existing text via synonym substitution or word order rearrangements. Advances in neural language models enable previously impractical generation-based techniques. We propose a steganography technique based on arithmetic coding with large-scale neural language models. We find that our approach can generate realistic looking cover sentences as evaluated by humans, while at the same time preserving security by matching the cover message distribution with the language model distribution.
%We find that the approach achieves close to theoretical bounds in terms of compression, and can fool human evaluators.
  %Unlike traditional cryptography which encrypts a secret message into an unintelligible form, steganography conceals the fact that communication is taking place at all by encoding a secret message into a natural-seeming carrier signal. In this work we propose \textit{generative word-choice steganography} in which a secret message is encoded in the choices a generative model for text makes during generation. To ensure that a third-party is unable to tell the difference between the encoded message and human written text, we leverage recent advancements in scaling large pretrained language models to formulate and approach this problem. We develop a theoretical framework which connects intuitions about entropy and information with the realities of steganography and use it to derive performance bounds. We propose specific tractable algorithms within this framework, and in experiments show the trade-off between words/bit and security and quantify the gap between our implementation and theoretical bounds.
\end{abstract}

\section{Introduction}

Cryptography is central to modern communication, but while it effectively conceals the content of a message it reveals that meaningful communication is taking place. Steganography answers an alternative question: how to conceal a message in some cover signal (an image, text etc.) such that an eavesdropper is not even aware any meaningful communication has taken place \cite{amin2003information,shirali2007text,kumar2010steganography}? Different from cryptography, in steganography security is derived from the inability to detect that a message exists within the cover signal, rather than the inability of an eavesdropper to determine the content of the message \cite{westfeld1999attacks}. 

%Given a fixed level of security, the goal of steganography is to maximize the efficiency of encoding.

Natural language is an especially useful cover signal for steganography because it is prevalent and innocuous in everyday life. Furthermore, unlike images or audio in which encoded information depends on exact pixel values, linguistic steganography (hiding information in choices of words) is untethered from any one medium. For example, the sender could encode a message on a computer, read off the cover text in person, and then the receiver could separately enter the cover text into her decoder and retrieve the message.

Linguistic steganography methods can be classified as either edit-based or generation-based \cite{bennett2004linguistic}. In edit-based methods, such as synonym substitution, the content of the cover text is selected by a human and slightly modified to encode information \cite{xiang2017novel,hefei2009steganalysis,xiang2018reversible}. In generative methods an entire block of text is generated while encoding the message reversibly in the choice of tokens. Traditionally, most practical stenography systems are edit-based. These methods only encode a small amount of information (for example, ~2 bits per tweet \cite{wilson2016avoiding}), whereas generation-based systems can encode an order of magnitude more information \cite{fang2017generating}. 

%For this reason, we focus on generation-based steganography methods in this paper.

Generation-based steganography has a well-established foundation in information theory \cite{cox2005information} with the aim of provably fooling any machine steganalysis. Due to weak language models practical performance at the scale of modern neural network-based models has not been considered or developed. One recent exception is a neural network-based approach using heuristic-based methods with the goal of fooling human eavesdroppers. These methods are theoretically sub-optimal, however, and while they demonstrate some fluency in generations the underlying language models are much worse than current state-of-the-art large models \cite{fang2017generating}.

In this paper we aim to get the best of both approaches with a linguistic steganography method based on arithmetic coding and modern language models. Our contribution is two-fold: 1) we show that a linguistic steganography approach combining arithmetic coding with state-of-the-art language models can achieve near-optimal statistical security; 2) human evaluations show that the cover text from our approach is able to fool humans even in the presence of context. Our code is available at \url{https://github.com/harvardnlp/NeuralSteganography}. A demo is available at \url{https://steganography.live/}

%In experiments probing the connection to theory and human evaluations we show that our approach both achieves better security guarantees than baselines, and successfully defends against human eavesdroppers. 

\section{\label{sec:ps}Background \& Related work}

 \begin{figure}[t]
\centering
\includegraphics[width=\linewidth]{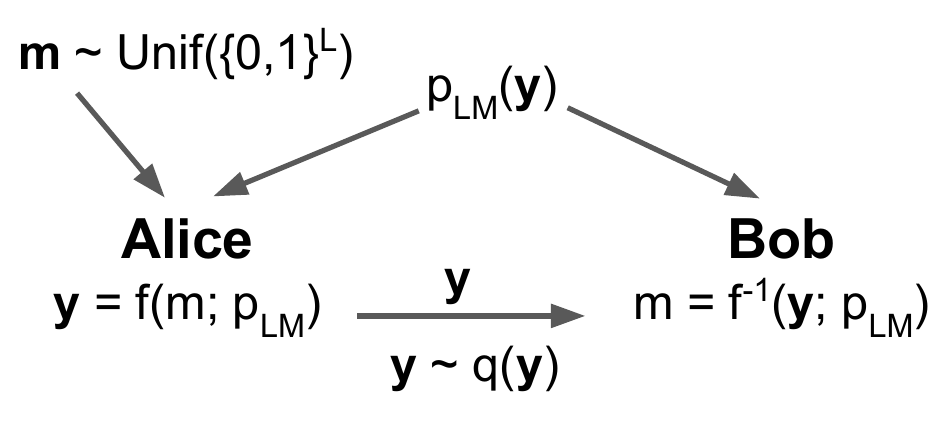}
\caption{\label{fig:problem} Problem setup. $\boldsymbol{m} \sim \text{Unif}(\{0,1\}^L)$ is the secret message, $\boldsymbol{y}$ is the cover text, $p(\boldsymbol{y})$ is the language model and $f$ is a deterministic invertible function. $f$ and the distribution of $m$ implicitly defines $q$.}
\vspace*{-0.3cm}
\end{figure}

\paragraph{Linguistic steganography}

We briefly give an overview of steganography as outlined in \cite{cox2005information} and diagrammed in Figure \ref{fig:problem}: Alice wants to communicate a hidden message $\boldsymbol{m} \sim \text{Unif}(\{0,1\}^L)$ with Bob by encoding it into a choice of natural language cover text $\boldsymbol{y}$. The uniform distribution is chosen for $\boldsymbol{m}$ without loss of generality: if $\boldsymbol{m}$ has additional structure it can be further compressed to a uniformly distributed random variable \cite{Han:2005:FSC:2263419.2271578}. Alice and Bob have both agreed on an invertible mapping $f$ which performs the steganography. Alice and Bob also both have access to the exact same language model, $p_{LM}(\boldsymbol{y})$, which $f$ can use during encoding and decoding. The steganography mapping $f$  and the language model $p_{LM}(\boldsymbol{y})$ form the key.

The combination of the distribution of $\boldsymbol{m}$ and deterministic function $\boldsymbol{y}=f(\boldsymbol{m})$ implicitly defines a distribution for $\boldsymbol{y}$ which we denote $q$. This is the cover distribution of natural language that an eavesdropper would observe. As described in \cite{cox2005information}, the security of the system is determined by $D_{KL}(q||P_{true})$ where $P_{true}$ is the true distribution of natural language. For example, if $D_{KL}=0$ then the system is perfectly secure because an eavesdropper could not distinguish the encoded messages from normal language. From an information theoretic perspective the security objective of steganography is therefore to ensure a small $D_{KL}$. At the same time, because $f$ is invertible the average number of bits that can be encoded is given by the entropy $H(q)$. Thus, the compression objective is to maximize $H(q)$.

\begin{figure}[t]
\centering
\includegraphics[width=1.02\linewidth]{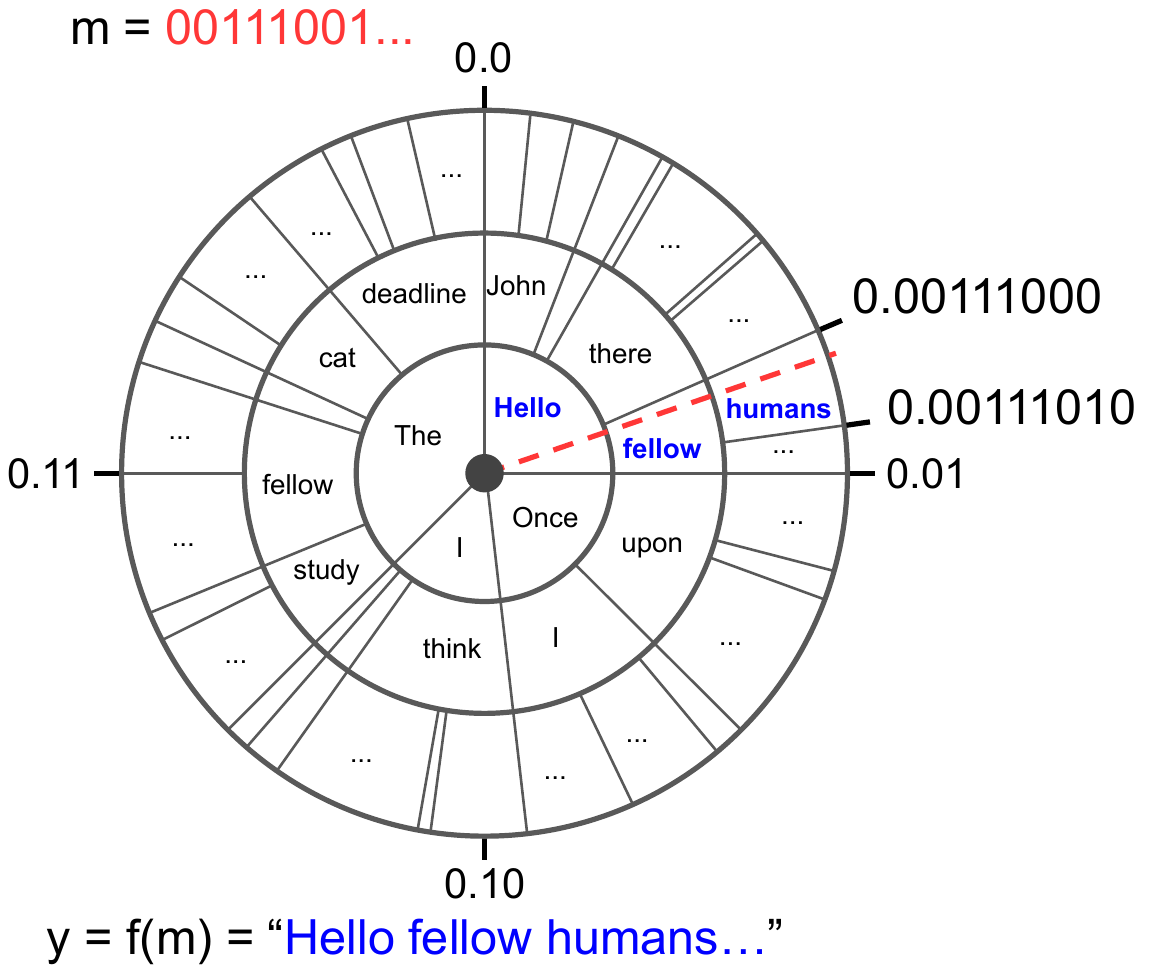}
\caption{\label{fig:arithmetic} Diagram of arithmetic coding for steganography. See Section \ref{sec:arithmetic} for details.}
\vspace*{-0.3cm}
%Starting at $t=1$ the fraction uniquely specifies an element based on which ``bin'' it falls in in the cumulative distribution function of $p_s(y_1)$. In this case it falls in the range of ``Hello'', so that word is output}
\end{figure}

\begin{table*}[t]
    \centering\footnotesize{
    \begin{tabular}{lp{13.5cm}}
    \toprule
    \textbf{Context} & kim jong il was the enigmatic leader of the most enigmatic country on earth. much about kim's life was, and is, shrouded in mystery. even the year of his birth -- 1942 -- has been contested by knowledgeable sources in south korea and russia, who claim kim was born a few years earlier but that his official birth date was pushed back to make him exactly 30 years younger than his father, north korea's founding leader, kim il sung. \\
    \midrule
    \midrule
    \textbf{Message 1} & Bob, I have something very important to tell you. Eve is not who she says she is. I know that you think she's your friend, but really she's a creepy stalker who wants to read all of your messages for some unknown reason. \\[0.3em]
    \textbf{Cover text 1} & That kim's birthdate was not even known to the people of south korea is even more baffling. however, there is another contradiction. bak tae-jung, the ruler of north korea at the time, claimed to have inherited kim's kim ju-ha-teh, his so-called golden heritage, from his father, the last kim Yun-jong. and he also \\
    \midrule
    \textbf{Message 2} & The secret code is X3SJ83F. Once you open the door you will see a staircase on your left and a hallway to your right. Go down the hallway and enter the code in the safe at the end of the hallway. \\[0.3em]
    \textbf{Cover text 2} & kim il sang, by the way, was not the first leader of north korea, but the first for nearly a thousand years.  his leader was kim jong il, the same person the people voted in to power to start with, without their knowledge or consent.  though it's been claimed by others that kim il sang fled north korea in 1940 to escape the reprisals, even that is disputed by k \\
    \bottomrule
    \end{tabular}
    }
    \caption{Steganography example. Two different encoded messages are produced given the same introductory context. The messages are first converted into bit strings and then mapped to cover text using the arithmetic steganography approach described in Section~\ref{sec:full_proc}.}
    \label{tab:example}
    \vspace*{-0.2cm}
\end{table*}

\paragraph{Generative coding techniques}

Recent work studying generative approaches for linguistic steganography have considered heuristic approaches to map uniform messages into similar distributions as natural language. \citet{fang2017generating} considers a block-based approach (\textbf{Block}), where the vocabulary is randomly split into $2^{|B|}$ bins and each bin is indexed by a $|B|$-bit string. Encoding is performed by splitting the message up into chunks $B_t$ and taking the maximum likelihood token as determined by the language model $p(y_t|y_{<t})$ that falls in bin $B_t$. \citet{yang2019rnn} proposes a related method based on constructing a Huffman coding (\textbf{Huffman}) at each step $t$ based on $p(y_t|y_{<t})$ and encoding the message in variable length chunks via the Huffman tree. Both objectives provide a ``quality'' parameter that trades off quality at the expense of encoding fewer bits.

These approaches target a different security objective than the information-theoretic view in \citet{cox2005information}. Instead of aiming to minimize the KL at maximal compression they aim to maximize generated quality at maximal compression \cite{yang2019rnn,fang2017generating}. 
% hey zack is that true? probably the easiest way  is to jsut delete this sentence, liek "The issue with these objectives are that they aim to fool human blablabla quality is most important
% go for it. I'm trying to make sure our submission (most recent) is going through
%ok, let me see what i can do
%As a proxy for generation quality they use the LL: $\E_q \log P_{true}(\boldsymbol{y})$. As $D_{KL}(q||P_{true})=H(q)-\E_q \log
%P_{true}(\boldsymbol{y})$ the difference in these two objectives is the $H(q)$ term. 
%The issue with this objective is that the KL identifies very low entropy $q$ distributions as insecure because an observer could over time determine that generations were not as varied as normal language, whereas 
%On the other hand, the LL objective identifies very low entropy $q$ distributions as the most secure because they produce likely natural language more frequently making it difficult for a human to identify that information is encoded. 
The issue with these approaches is that they aim to fool human eavesdroppers for whom generation quality is most important; but are susceptible to machine-based eavesdroppers which can in principle identify statistical patterns in generations. 

Concurrent with this work, \citet{dai-cai-2019-towards} conduct a related analysis in terms of KL with a modified Huffman algorithm. Experiments consider the distribution of KL values, although no human evaluation is performed.

%In this work we investigate both objectives.

%A generative method can also be designed to ensure fluency and security, but extra steps must be taken to ensure that the content of the generated carrier message makes sense in context. If the generated text sounds and looks like natural language but is completely out of context then the steganography will be ineffective. While this is a downside of unconditional generative steganography, it is not hard to imagine practical implementations that fix this problem. For example, consider a language generation model trained as a chat bot for dialogue. The sender could initiate a conversation with the first part of a message encoded in a carrier of a greeting. The sender could receive the message chunk and reply with text that makes sense given the initial message. The sender would then encode the next message chunk in a response to the receiver by conditioning the dialogue model on all past text. In this way the sender could send the message in chunks to the receiver with a generative system, while to an eavesdropper the communication seems like a mundane discussion about the weather or family. Compared to other simpler forms of steganography, such as using code words \cite{kahn1996history}, this method allows one to encode arbitrary bit sequences relatively efficiently.

\section{\label{sec:arithmetic}Arithmetic coding}

Arithmetic coding is a data compression method designed specifically to code strings of elements with a known probability distribution \cite{rissanen1979arithmetic,zoph2015much}. For long strings the coding is optimal; it compresses information to its entropy \cite{rissanen1979arithmetic}. In practice, it is often more efficient than Huffman coding because it does not require blocking. Arithmetic coding traditionally maps a string of elements to a uniformly distributed binary string. To use such a coding for steganography we reverse the order: first a (uniformly sampled) message is selected, then the message is mapped to a sequence (words).

The coding scheme is demonstrated in Figure \ref{fig:arithmetic}. In this work the probability distribution comes from the conditional distributions of a pretrained language model, but for illustration purposes a hand-crafted example distribution is used in the diagram. Concentric circles represent timesteps; the innermost represents $t=1$, the middle $t=2$, and the outer $t=3$. Each circle represents the conditional distribution $p(y_t|y_{<t})$. For example, given that $y_1=\text{``Once''}$, $p(y_2|y_1)$ has ``upon'' and ``I'' as the only possible tokens with equal probability. The circle diagram spans [0,1) from 0 at the top, clockwise around to 1. 

To encode the message into text, the secret message $m$ is viewed as a binary representation of a fraction in the range $[0,1)$. This fraction uniquely marks a point on the edge of the circle, as well as a line from the origin to the point. Encoding is performed by simply reading off the tokens corresponding to the bins. Encoding stops when the list of tokens unambiguously defines the message.

Decoding is performed via the reverse operation: the sequence of natural language tokens progressively narrows the range of possible messages. Assuming that the original message is encoded with a predetermined end token, decoding terminates once both sides of the range of possible messages includes the end token.

\citet{sallee} show that just as arithmetic coding is optimal for data compression, it is also optimal for steganography. Specifically, given any goal distribution $p_s(\boldsymbol{y})$ that one wants to sample from, starting with a uniform $\boldsymbol{m}$ and applying the deterministic steganography procedures yields a distribution $q = p_s$ or equivalently $D_{KL}(q||p_s)=0$ for long sequences. This further ensures that $H(q)=H(p_s)$ and therefore the number of bits encoded on average is equal to the entropy of $p_s$. Intuitively, this works because higher probability sequences map to larger ``chunks'' of the pie and therefore require fewer bits to uniquely determine.

In this work we apply arithmetic coding to linguistic steganography. Given a pretrained language model, this approach can be directly applied to produce cover texts given some preliminary context. To ensure that these texts seem more natural to human eavesdroppers we introduce two parameters to trade off quality for compression: we modulate the LM distribution by a temperature $\tau$, and we truncate the distribution to a top-$k$ tokens at each position to ensure generations do not include tokens from the long tail.

In practice, a naive implementation of arithmetic coding quickly runs into precision limitations. We use a common equivalent variant based on fixed precision binary fractions \cite{rubin1979arithmetic}.

\section{Experiments and results}
\begin{figure}[!t]
        \centering
        \includegraphics[width=0.97\linewidth]{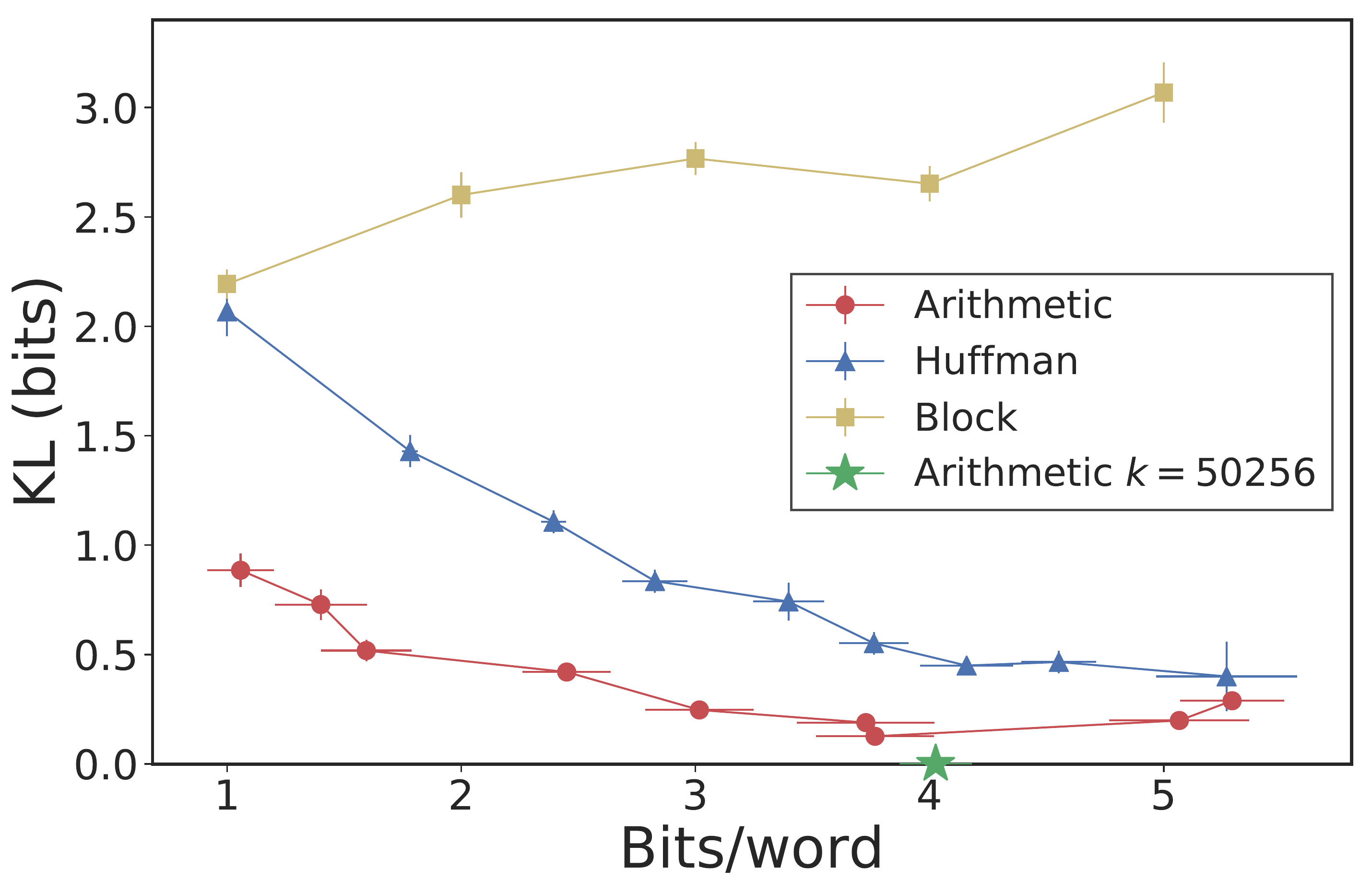}
        \caption{\label{fig:auto} Information theoretic evaluation. KL and bits/word are evaluated for each algorithm at different values of the tuning parameter; each datapoint gives the mean and standard error for repeated samples with a fixed value of the tuning parameter. We show the standard error bars for both bits/word and KL.}
        \vspace*{-0.2cm}
\end{figure}
\subsection{Experimental Setup} We use the 345M parameter GPT-2 model as our language model $p_{LM}$ \cite{radford2019language}. We run our experiments on the CNN/Dailymail (CNNDM) dataset \cite{hermann2015teaching,nallapati2016abstractive}, which is used to provide  context for generation\footnote{Conditional generation makes it more challenging to fool humans and is thus better for comparisons. CNNDM presents an additional challenge as poor generations may have clear factual errors.}. We take the first three sentences of each news article as the context on which to condition our language model, and use the indicated steganography algorithm to generate an entire sentence for estimation or evaluation given a uniform random message. We compare the proposed arithmetic coding-based algorithm with the Block \cite{fang2017generating} and Huffman \cite{yang2019rnn} methods as baselines. The baselines are re-implemented to allow for a fair comparison using the strong GPT-2 language model. Each method has a parameter that controls the various tradeoffs: for arithmetic encoding we vary the temperature from 0.4 to 1.2 with $k=300$, for Huffman encoding we vary the truncation length from $2^1$ to $2^8$, for block encoding we vary the block size from $2^1$ to $2^5$. For the KL analysis we also evaluate arithmetic coding with $\tau=1$ and $k=50256$, i.e. the unmodulated language model.

\subsection{Metrics}

We consider several different metrics to assess performance. 
In all cases, it is important that the algorithms encode information efficiently. The  metric \textbf{Bits/word} is the ratio of message bits encoded to sentence length.

Another key aspect of the system is being able to fool a statistical adversary in a \textit{distributional} sense, i.e. when the adversary gets multiple samples from $q$. This can be quantified by the \textbf{KL} divergence between $q$ and $P_{true}$. Since we do not have access to $P_{true}$ we resort to $D_{KL}(q||p_{LM})$. $p_{LM}$ might not match the distribution of natural language, but it allows for relative comparisons of steganography algorithms. The remaining gap is a function of language model performance, which is not the focus of this work. For each model the KL is estimated via MC sampling on the news data.

%The KL experiments probe the ability of an automated system to discover patterns that distinguish the generations from natural language. In practice, we are additionally concerned with the ability of humans to identify single out-of-place sentences. 
Finally it is important to be able to convince a \textbf{human adversary} that the text is natural. This differs from statistical considerations because a human will only see a single sample from $q$ at a time. To this end, we conduct a human study aligned with the information theoretic evaluation above, using Amazon Mechanical Turk. For each of the three methods we generate sentences with randomly sampled messages conditioned on a news article context, and ask participants to indicate if the sentence is a likely next sentence in context. We repeat this experiment with the true next sentence as well. See the Supplementary Materials for more details.

\begin{figure}[!t]
        \centering
        \includegraphics[width=0.97\linewidth]{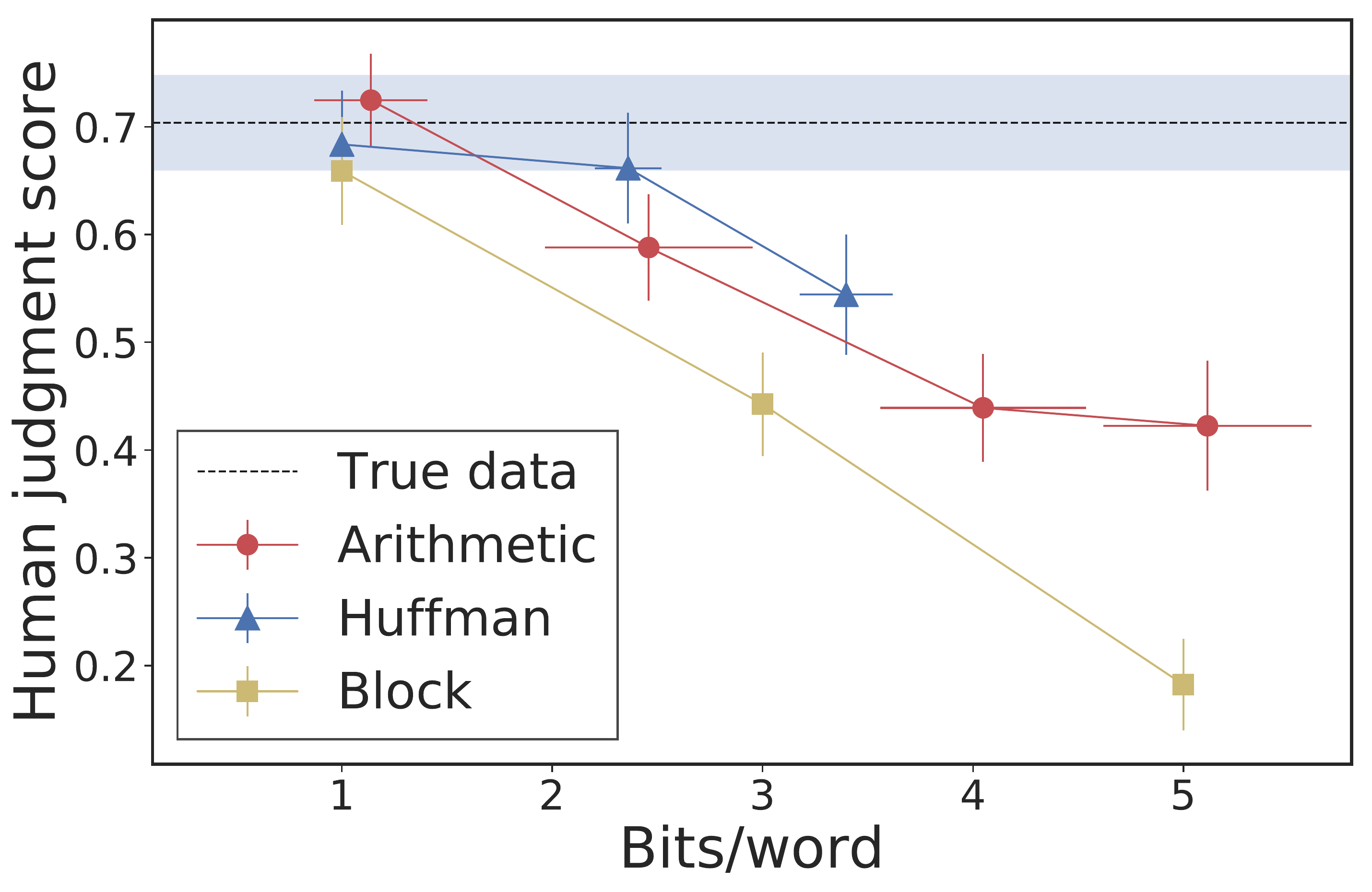}
        \caption{\label{fig:human}Human evaluation. Y axis shows the percentage of humans considering the generation to be a proper continuation of the context. Huffman coding achieves the best generation quality as judged by humans, while arithmetic coding is comparable (the difference is not statistically significant).}
        \vspace*{-0.3cm}
\end{figure}

\subsection{Quantitative results}
Figure~\ref{fig:auto} shows the information theoretic evaluation. For any bits/word in the range of (1, 5), arithmetic coding achieves the lowest KL, i.e. the distribution of the encoded texts is the most similar to the language model distribution, providing the best defensive against algorithmic detections. Most striking, arithmetic coding with the unmodulated language model induces a $q$ distribution with a KL of 4e-8 nats. This indicates that, consistent with theory \cite{sallee}, arithmetic coding enables generative steganography matching the \textit{exact} distribution of the language model used. The small but positive KL value for arithmetic coding with $k=300$ comes from the slight distributional difference when the long tail is truncated.

Figure~\ref{fig:human} shows the human evaluation results. Within 3 bits/word both Huffman and arithmetic coding give plausible next sentences over 50\% of the time, and at 1 bit/word both methods lead to cover texts statistically indistinguishable from human-written sentences. Although Huffman coding reaches slightly better performance than arithmetic coding, the transferred cover text distribution is statistically different from natural text (see Figure~\ref{fig:auto}), which can be potentially discovered by automatic systems. Compared to the KL analysis which focuses on the relative performance of the algorithms, the human evaluation highlights the realizable generation quality enabled specifically by large pretrained language models. Still, there is room for improvement in terms of language modeling capabilities: whereas the arithmetic approach reaches minimum KL around 4 bits/word with $\tau=1.0$, only at 1 bit/word with $\tau=0.4$ do the generations fool humans. 

\subsection{Qualitative results}
\label{sec:full_proc}
Steganography fundamentally deals with hiding uniformly random message bits in cover text. In many cases, however, the message may itself be natural language. In these cases we have an additional choice of how the text is converted to bits. One option is simply the unicode encoding (i.e. UTF-8), though this is highly inefficient. Instead, we can use the same arithmetic coding algorithm in reverse, with an empty context, to compress the message text into near-uniformly distributed bits \cite{Han:2005:FSC:2263419.2271578}. The full procedure is as follows:

\begin{enumerate}
    \item Alice decodes message text into bits via arithmetic coding with an empty context.
    \item Alice encodes bits into cover text via arithmetic coding with a predetermined context.
    \item Alice sends cover text over public channel.
    \item Bob decodes cover text into bits via arithmetic coding with the predetermined context.
    \item Bob encodes bits into message text via arithmetic coding with empty context.
\end{enumerate}

\noindent
Because the same strong LM is used with different contexts for the two applications of arithmetic coding, the message text length and cover text length will be comparable (modulo a small overhead) but the content of the text can be completely different. This is demonstrated in Table~\ref{tab:example}, which shows two examples of natural language messages (Step 1) encoded in unrelated and innocuous cover text (Step 3). In both cases the message is encoded efficiently into cover text which is largely fluent and coherent, and follows from the given context.

\section{Conclusion}
We demonstrate how the combination of large pretrained language models and an arithmetic coding-based steganography algorithm allows for a steganography system that can fool both statistical adversaries who analyze the full generative distribution, and human adversaries who analyze a single sample. While our work shows the potential for high-quality efficient steganography and the realizable optimality of arithmetic coding, future advancements in language modeling can push steganographic performance even further. With current state-of-the-art language models the steganography algorithms studied generate most convincing cover texts at low compression, where the KL is moderate. As language models continue to improve, they can be directly plugged into our arithmetic approach to maximize steganographic performance.

%The quality of generated text is approaching the level of human text \cite{hassan2018achieving}.
%We leverage advances in language models to map bit strings to a sequence of words. To ensure the security, we minimize KL, while for efficiency we maximize information carried by each word. 
%Arithmetic encoding provides the near-optimal bits/word when we require the encoded distribution be the same as the language model distribution \cite{saniei2013security,moffat1989word}. 
%In practice, we get a trade-off between compression and security by using arithmetic coding. Our approach, combined with modern powerful language models, is able to generate realistic looking sentences and fool humans even in the presence of context, while at the same time the encoded text distribution is much closer to the language model distribution compared to existing methods.

%However, we do note that one limitation of our approach is that we do not consider how secrecy can be preserved in the presence of an eavesdropper with access to the exact language model being used and knowledge that secret communication is occurring. While outside the considerations of steganography, it is an important practical question that we leave for future work. Overall, we believe that our approach represents an initial step towards generative steganography leveraging large pretrained language models.

\newpage
\section*{Acknowledgments}
This work was supported by NSF 1845664. YD was supported 
by a Baidu AI Fellowship. 

\bibliography{emnlp2018}
\bibliographystyle{acl_natbib_nourl}

\newpage

\appendix
\vspace{-3in}
\section{Human evaluation details}

The human evaluation uses the same data as the KL evaluation, consisting of 3 sentence contexts and 1 sentence generations. The generations come from 11 sources: arithmetic coding with temperatures 0.4, 0.7, 1.0, 1.2, Huffman encoding with truncation lengths $2^1$, $2^3$, $2^5$, block encoding with block sizes $2^1$, $2^3$, $2^5$, and the original human-written next sentences. For the steganography systems the message is sampled uniformly at random. 80 generations from each source are gathered and randomly mixed into HITs with 5 generations per HIT.

Each HIT asks, ``Given the start of a news article: $<$article$>$ Is the following a likely next sentence: $<$generation$>$ Yes/No. Please consider if it makes sense in the context of the article, if it contains factually correct information, and if it is grammatically correct.'' The workers are additionally given further instructions defining what it means to make sense in context, contain factual information, and be grammatically correct. Human judgment is calculated from the mean response of the workers. We measure absolute human judgment instead of a head-to-head comparison because it is better aligned with the real world use: a viewer would not have access to two continuations of a paragraph, they would have to decide if the given text alone feels out of place.

In addition, each HIT includes one attention check question that can only be passed if the worker reads the entire article closely. HITs with failed attention checks, HITs from workers with frequent failed attention checks, and HITs completed too quickly relative to the other workers are not used for analysis.

\section{Additional Examples}

Tables \ref{tab:arithmetic}, \ref{tab:huffman} and \ref{tab:block} gives additional examples of the different steganography algorithms. For each, the context (human written) comes from the first three sentences of a randomly selected CNN/DM article. The message is a uniformly sampled bit string, and the generated cover text is cut off after one sentence. The steganography parameters used for the three tables are $\tau=0.7$, $k=300$ for arithmetic, $\text{truncation length}=2^3$ for Huffman, and $\text{block size}=2^3$ for block, which give similar compression efficiencies. The examples are the first four generated for each model and are not curated. For more samples and samples from models with other steganography parameters see \url{https://github.com/harvardnlp/NeuralSteganography}.

\begin{table*}[t]
    \centering\footnotesize{
    \begin{tabular}{lp{13.7cm}}
    \toprule
    \multicolumn{2}{l}{\textbf{Arithmetic}, $\tau=0.7$, $k=300$} \\
    \midrule
\textbf{Context} & kathmandu, nepal -lrb- cnn -rrb- rescue crews and residents in nepal early sunday began the desperate search for survivors after a magnitude-7 .8 quake near the capital of kathmandu a day earlier flattened homes, buildings and temples, causing widespread damage across the region and killing more than 1,800 people. follow the latest coverage of nepal earthquake. \\[0.3em]
\textbf{Cover text} &  While rescue efforts had been concentrated in the capital, rescue operations continued across the entire country, with rescuers unable to find anyone alive in any of many districts. \\
\midrule
\textbf{Context} & a man was caught allegedly trying to smuggle two pounds of cocaine worth \$30,000 in pairs of sneakers at jfk airport earlier this month. on april 7, thenga adams, flying from guyana in south america was arrested after customs at jfk in new york searched the sneakers in his luggage. when customs opened the soles of the athletic shoes they found \$30,000 worth of cocaine, say airport officials. \\[0.3em]
\textbf{Cover text} &  the shoes were later returned to the airport and it is unknown if the shoes were recovered from the plane or not. \\
\midrule
\textbf{Context} & a talented body painter is travelling across america merging members of the public into beautiful backgrounds. natalie fletcher's' 100 bodies across america 'project sees painted individuals blended into the likes of tourist hotspots, forests and ruins. for each work natalie travels to a certain spot in the country before selecting volunteers - not models - who she then paints. \\[0.3em]
\textbf{Cover text} &  to add to the realism of the process, the volunteers are drugged and disoriented after their arrival, and the final scenes are constructed from a series of real-life photographs taken by those who have tried to walk across America. \\
\midrule
\textbf{Context} & -lrb- cnn -rrb- sarah brady, who with her husband, james brady, pushed for stricter gun control laws, including the brady handgun violence prevention act, died friday, her family said. brady, 73, died of pneumonia, the family said. "sarah courageously stepped up after jim was shot to prevent others from enduring what our family has gone through, and her work has saved countless lives," their statement said. \\[0.3em]
\textbf{Cover text} &   The brady handgun violence prevention act, or HRPA, bans the sale and possession of handguns with a magazine capacity greater than 10 rounds. \\
    \bottomrule
    \end{tabular}
    }
    \caption{Additional steganography examples using the arithmetic method. Messages are uniformly random bits, the context is the first three sentences from CNN/DM articles and the generated cover text is cut off after one sentence.}
    \label{tab:arithmetic}
\end{table*}

\begin{table*}[t]
    \centering\footnotesize{
    \begin{tabular}{lp{13.7cm}}
    \toprule
    \multicolumn{2}{l}{\textbf{Huffman}, $\text{truncation length}=2^3$} \\
    \midrule
\textbf{Context} & frank ernest shepherd, iii of houston, texas, led police on a high-speed chase after being pulled over for a traffic stop that ended when he slammed into another vehicle. an unarmed man shot dead by police on live television on wednesday afternoon was a 41-year-old father of three who had another baby on the way. frank ernest shepherd, iii of houston, texas, led police on a high-speed chase after being pulled over for a traffic stop that ended when he slammed into another vehicle. \\[0.3em]
\textbf{Cover text} &  a 41 of 33 (82\%) voters in the state of america believe the police should always be on call and armed, while only 22\% think they are under the same authority to make arrest and only 16\% of americana believe the state should not use deadly force to defend itself. \\
\midrule
\textbf{Context} & -lrb- cnn -rrb- ahmed farouq didn't have the prestige of fellow al qaeda figure osama bin laden, the influence of anwar al-awlaki, or the notoriety of adam gadahn. still, he was a big deal. that's the assessment of multiple sources on a man who may not have been well-known in the west, but nonetheless had a special role in the terrorist group. \\[0.3em]
\textbf{Cover text} &  In addition, there was an al Qaeda operative named Abu al-Khalid al Qahtaniai, also called Abu Musab Abu al-Khatib, or Abu Muhaysin al Qahtan, born in Yemen. \\
\midrule
\textbf{Context} & the state of oregon on friday released 94,000 emails involving the fiancee of former gov. john kitzhaber, who resigned amid scandal earlier this year over allegations that the former first lady used her relationship with him to land contracts for her business. the emails show the very active role that cylvia hayes, kitzhaber's longtime companion, played in his administration, as well as the tension that it sometimes raised with the governor's staff. \\[0.3em]
\textbf{Cover text} &  hayes has also been linked to the former president, and was involved as the secretary of state's chief deputy, during her first term as secretary in 2006. \\
\midrule
\textbf{Context} & masters wednesday began on a sombre note when australian marc leishman, who got so close to victory in 2013 before finishing tied fourth behind countryman adam scott, had to withdraw. his wife audrey spent part of last week in an induced coma owing to a serious infection. leishman left augusta taking with him the best wishes of his sport. \\[0.3em]
\textbf{Cover text} &  he said he had been "very disappointed in myself and my team for the last few days", but added:  "I'm happy with my team and I'm happy with the fact my race went well in a very difficult condition. \\
    \bottomrule
    \end{tabular}
    }
    \caption{Additional steganography examples using the Huffman method. Messages are uniformly random bits, the context is the first three sentences from CNN/DM articles and the generated cover text is cut off after one sentence.}
    \label{tab:huffman}
\end{table*}

\begin{table*}[t]
    \centering\footnotesize{
    \begin{tabular}{lp{13.7cm}}
    \toprule
    \multicolumn{2}{l}{\textbf{Block}, $\text{block size}=2^3$} \\
    \midrule
\textbf{Context} & the ten houses in australia that won't sell, despite being heavily discounted, have been revealed. a recent report, released by sqm research, found the properties across the country which haven't been able to spark enough interest among home buyers even after being on the market for up to two years. among them includes a one-bedroom home in oberon, central west of nsw, which was listed 800 days ago for \$299,000 but the price has since been cut by 50 per cent, the property observer reports. \\[0.3em]
\textbf{Cover text} &  this particular piece sits empty for now, with an estimate from SQMR at just shy, 1,500\$ the total value is \$716M. \\
\midrule
\textbf{Context} & doctors have been accused of letting a woman die after wrongly concluding she had terminal cancer. medics told the family of margaret hesketh that there was little they could do for the 70-year-old because she was riddled with tumours, her children will claim at an inquest this week. the family claim she was then put on a discredited liverpool care pathway-style treatment for the dying -- having fluid and nutrition drips removed six days before she passed away. \\[0.3em]
\textbf{Cover text} &  her cancer returned when hospital tests revealed tumores – but when hospital tests turned on the liquid to cause further harm when her body rejected it she went into a vegetas mode and began suffering seizures at least twice daily over weeks and then went back on this luke-full path to cancer again and began suffering another seizures a week in and weeks – which continued for months and was then confirmed after she finally was put in charge of the intensive rehabilitation treatment she had begun – and that doctors were now claiming the fluid to treat tumor cells, was 'an important diagnostic tool for people' when really it had caused a death -- her family claims her own doctors admitted she might be terminomised but her family has now spoken about the false claims at what the hearing today was set at by Sir James Gaventa and Sir Peter Fahlesworth, the medical directors who are investigating and looking further for evidence of misconduct by they medis. \\
\midrule
\textbf{Context} & children may be drawn to greasy fries and burgers, but fatty foods can reduce the speed at which their brains work, according to a new study. it found that children who ate a diet higher in saturated fats and cholesterol had slower reaction times and a poorer working memory. children who ate the fatty diet performed worse when they were given a task-switching game to complete, the researchers said. \\[0.3em]
\textbf{Cover text} &  The results suggest the fats may interfere not only by blocking fat transport between various locations but actually slowing it along. \\
\midrule
\textbf{Context} & bruce jenner could be sued by the stepchildren of the woman who died in the pacific coast highway car crash earlier this year - despite the fact they had `virtually no relationship '. kim howe's two adult stepchildren are said to be considering suing the former olympic athlete turned reality television star following her death. mrs howe was driving a white lexus, which jenner's cadillac escalade rear-ended on a segment of the pacific coast highway in malibu, california, on february 7. \\[0.3em]
\textbf{Cover text} &  the accident led the authorities within days after her fatal, on March 13. \\
    \bottomrule
    \end{tabular}
    }
    \caption{Additional steganography examples using the block method. Messages are uniformly random bits, the context is the first three sentences from CNN/DM articles and the generated cover text is cut off after one sentence.}
    \label{tab:block}
\end{table*}

\end{document}

%% file: preamble.tex
%% Courtesy: Daniel Spielman, via Madhu Sudan --> Vinod Vaikuntanathan --> Aloni Cohen

%--------------
%% preamble.tex
%% this should be included with a command like
%% \input{preamble.tex}
%% \lecture{1}{September 4, 1996 }{Daniel A. Spielman}{name
%%  of poor scribe}

\hbadness=10000
\vbadness=10000

\newcommand{\handout}[5]{
   \renewcommand{\thepage}{#1-\arabic{page}}
   \noindent
   \begin{center}
   \framebox{
      \vbox{
    \hbox to 5.78in { {\bf #1}
     	 \hfill #2 }
       \vspace{4mm}
       \hbox to 5.78in { {\Large \hfill #5  \hfill} }
       \vspace{2mm}
       \hbox to 5.78in { {\it #3 \hfill #4} }
      }
   }
   \end{center}
   \vspace*{4mm}
}

\newcommand{\lecture}[4]{\handout{#1}{#2}{Lecturer:
#3}{Scribe: #4}{Lecture #1}}

\newtheorem{theorem}{Theorem}
\newtheorem{corollary}[theorem]{Corollary}
\newtheorem{lemma}[theorem]{Lemma}
\newtheorem{observation}[theorem]{Observation}
\newtheorem{proposition}[theorem]{Proposition}
\newtheorem{definition}[theorem]{Definition}
\newtheorem{claim}[theorem]{Claim}
\newtheorem{fact}[theorem]{Fact}
\newtheorem{assumption}[theorem]{Assumption}
\newtheorem{exercise}[theorem]{Exercise}

%%%% GENERAL COMMANDS %%%%%%%%%%%%%%%%%%%%%%%%%%%%%%%%%

\newcommand{\dis}{\mathop{\mbox{\rm d}}\nolimits}
\newcommand{\per}{\mathop{\mbox{\rm per}}\nolimits}
\newcommand{\area}{\mathop{\mbox{\rm area}}\nolimits}
\newcommand{\cw}{\mathop{\rm cw}\nolimits}
\newcommand{\ccw}{\mathop{\rm ccw}\nolimits}
\newcommand{\DIST}{\mathop{\mbox{\rm DIST}}\nolimits}
\newcommand{\OP}{\mathop{\mbox{\it OP}}\nolimits}
\newcommand{\OPprime}{\mathop{\mbox{\it OP}^{\,\prime}}\nolimits}
\newcommand{\ihat}{\hat{\imath}}
\newcommand{\jhat}{\hat{\jmath}}
\newcommand{\abs}[1]{\mathify{\left| #1 \right|}}

%%%%% PROOF ENVIRONMENTS %%%%%%%%%%%%%%%%%%%%%%%%%%%%%%%%%%%%%%%%%%%%%%%%%%%%%%%%%%%%

\newenvironment{proof-sketch}{\noindent{\bf Sketch of Proof}\hspace*{1em}}{\qed\bigskip}
\newenvironment{proof-idea}{\noindent{\bf Proof Idea}\hspace*{1em}}{\qed\bigskip}
\newenvironment{proof-of-lemma}[1]{\noindent{\bf Proof of Lemma #1}\hspace*{1em}}{\qed\bigskip}
\newenvironment{proof-attempt}{\noindent{\bf Proof Attempt}\hspace*{1em}}{\qed\bigskip}
\newenvironment{proofof}[1]{\noindent{\bf Proof}
of #1:\hspace*{1em}}{\qed\bigskip}
\newenvironment{remark}{\noindent{\bf Remark}\hspace*{1em}}{\bigskip}

% \makeatletter
% \@addtoreset{figure}{section}
% \@addtoreset{table}{section}
% \@addtoreset{equation}{section}
% \makeatother

\newcommand{\FOR}{{\bf for}}
\newcommand{\TO}{{\bf to}}
\newcommand{\DO}{{\bf do}}
\newcommand{\WHILE}{{\bf while}}
\newcommand{\AND}{{\bf and}}
\newcommand{\IF}{{\bf if}}
\newcommand{\THEN}{{\bf then}}
\newcommand{\ELSE}{{\bf else}}

\makeatletter
\def\fnum@figure{{\bf Figure \thefigure}}
\def\fnum@table{{\bf Table \thetable}}
\long\def\@mycaption#1[#2]#3{\addcontentsline{\csname
  ext@#1\endcsname}{#1}{\protect\numberline{\csname
  the#1\endcsname}{\ignorespaces #2}}\par
  \begingroup
    \@parboxrestore
    \small
    \@makecaption{\csname fnum@#1\endcsname}{\ignorespaces #3}\par
  \endgroup}
\def\mycaption{\refstepcounter\@captype \@dblarg{\@mycaption\@captype}}
\makeatother

\newcommand{\figcaption}[1]{\mycaption[]{#1}}
\newcommand{\tabcaption}[1]{\mycaption[]{#1}}
\newcommand{\head}[1]{\chapter[Lecture \##1]{}}
\newcommand{\mathify}[1]{\ifmmode{#1}\else\mbox{$#1$}\fi}
\newcommand{\bigO}O
\newcommand{\set}[1]{\mathify{\left\{ #1 \right\}}}
\def\half{\frac{1}{2}}

% Command that ignores input.
\newcommand{\remove}[1]{}
\newcommand{\ignore}[1]{}
\newenvironment{ignoreme}{\ignore{}{}}

%%%%%%%%%%%%%% LATTICES %%%%%%%%%%%%%%%%%%%%%%%%%%%%%%%%%%%%%%%%%%%%%%%%%%%%%%%%%

\def\hh{\sigma^{\text{\rm\tiny top}}}
\def\ph{\sigma^{\text{\rm\tiny bot}}}
\def\hM{M^{\text{\rm\tiny top}}}
\def\pM{M^{\text{\rm\tiny bot}}}
\def\lsb{\mathsf{lsb}}

\newcommand{\gapSVP}{\mathsf{gapSVP}}
\newcommand{\SIVP}{\mathsf{SIVP}}

\def\dlwe{\mathsf{DLWE}}

\def\Zp{\Z_p}
\newcommand{\ip}[1]{\langle #1 \rangle}
\def\Psibar{\overline{\Psi}}

%Security Parameter
\newcommand{\secparam}{\kappa}
\newcommand{\secp}{\secparam}

\newcommand{\svp}{\mathsf{SVP}}

% Vectors, Matrices and such

\def\veca{\vc{a}}
\def\vecb{\vc{b}}
\def\vecc{\vc{c}}
\def\vecd{\vc{d}}
\def\vece{\vc{e}}
\def\vecm{\vc{m}}
\def\vecs{\vc{s}}
\def\vect{\vc{t}}
\def\vecv{\vc{v}}
\def\vecx{\vc{x}}
\def\vecy{\vc{y}}

\def\Z{\mathbb{Z}}
\def\R{\mathbb{R}}
\def\Q{\mathbb{Q}}

% Changing QED symbol in claim proofs
\newenvironment{claimproof}{\begin{proof}
\renewcommand{\qedsymbol}{{$\blacksquare$}}
}{\end{proof}}

% Calligraphic and blackboard type letters.

\def\cA{{\cal A}}
\def\cB{{\cal B}}
\def\cC{{\cal C}}
\def\cD{{\cal D}}
\def\cE{{\cal E}}
\def\cF{{\cal F}}
\def\cG{{\cal G}}
\def\cH{{\cal H}}
\def\cI{{\cal I}}
\def\cJ{{\cal J}}
\def\cK{{\cal K}}
\def\cL{{\cal L}}
\def\cM{{\cal M}}
\def\cN{{\cal N}}
\def\cO{{\cal O}}
\def\cP{{\cal P}}
\def\cQ{{\cal Q}}
\def\cR{{\cal R}}
\def\cS{{\cal S}}
\def\cT{{\cal T}}
\def\cU{{\cal U}}
\def\cV{{\cal V}}
\def\cW{{\cal W}}
\def\cX{{\cal X}}
\def\cY{{\cal Y}}
\def\cZ{{\cal Z}}
%%%%%%%%%%%%%%%%%
\def\bbC{{\mathbb C}}
\def\bbE{{\mathbb E}}
\def\bbF{{\mathbb F}}
\def\bbG{{\mathbb G}}
\def\bbM{{\mathbb M}}
\def\bbN{{\mathbb N}}
\def\bbQ{{\mathbb Q}}
\def\bbR{{\mathbb R}}
\def\bbV{{\mathbb V}}
\def\bbZ{{\mathbb Z}}

\def\Zq{\bbZ_q}

%%%%%%%%%%%%%%%%%

% Rounding commands

\newcommand{\ceil}[1]{\left\lceil #1 \right\rceil}
\newcommand{\floor}[1]{\left\lfloor #1 \right\rfloor}
\newcommand{\round}[1]{\left\lfloor #1 \right\rceil}

% Other short-hands

\def\binset{\{0,1\}}
\def\pmset{\{\pm 1\}}
\def\ind{\mathbbm{1}}

\newcommand{\norm}[1]{\left\| {#1} \right\|}
\newcommand{\norminf}[1]{\left\| {#1} \right\|_{\infty}}

% Assignments
\def\getsr{\stackrel{\scriptscriptstyle{\$}}{\gets}}
\def\getsd{{:=}}
\def\bydef{\triangleq}
\def\getsf{{\gets}}

% Asymptotics

\def\poly{{\rm poly}}
\def\polylog{{\rm polylog}}
\def\polyloglog{{\rm polyloglog}}
\def\negl{{\rm negl}}
\newcommand{\ppt}{\mbox{{\sc ppt}}}
\def\Otilde{\widetilde{O}}

% Indistinguishability
\newcommand{\cind}{{\ \stackrel{c}{\approx}\ }}
\newcommand{\sind}{{\ \stackrel{s}{\approx}\ }}

% Complexity classes

\def\NP{\mathbf{NP}}
\def\Ppoly{{\mathbf{P}/\poly}}

% Cryptographic assumptions

\newcommand{\ddh}{\mathrm{DDH}}
\newcommand{\cdh}{\mathrm{CDH}}
\newcommand{\dlin}{\text{\rm $d$LIN}}
\newcommand{\lin}{\text{\rm Lin}}
\newcommand{\sxdh}{\mathrm{SXDH}}
\newcommand{\rsa}{\mathrm{RSA}}
\newcommand{\sis}{\mathrm{SIS}}
\newcommand{\isis}{\mathrm{ISIS}}
\newcommand{\lwe}{\mathsf{LWE}}
\newcommand{\qr}{\mathrm{QR}}

% Types of attacks

\newcommand{\adv}{\mathrm{Adv}}
\newcommand{\dst}{\mathrm{Dist}}
\newcommand{\leak}{\mathrm{Leak}}
\newcommand{\forge}{\mathrm{Forge}}
\newcommand{\col}{\mathrm{Col}}
\newcommand{\invt}{\mathrm{Inv}}
\newcommand{\cpa}{\text{\rm CPA}}
\newcommand{\kdm}{\mathrm{KDM}}
\newcommand{\kdi}{\mathrm{KDM}^{(1)}}
\newcommand{\kdmn}{\mathrm{KDM}^{(\usr)}}
\newcommand{\ibe}{\mathrm{IBE}}
\newcommand{\good}{\mathrm{GOOD}}
\newcommand{\legal}{\mathrm{L}}

% Quadratic residuosity related

\newcommand{\qrs}{\mathbb{QR}}
\newcommand{\js}{\mathbb{J}}

% Linear algebra

\newcommand{\mx}[1]{\mathbf{{#1}}}
\newcommand{\vc}[1]{\mathbf{{#1}}}
\newcommand{\gvc}[1]{\bm{{#1}}}

\newcommand{\rk}{\text{\rm Rk}}
\newcommand{\spn}{\text{\rm Span}}

% Probability

\newcommand{\Ex}{\mathop{\bbE}}
\newcommand{\cov}{\mathop{\text{\rm Cov}}}

\newcommand{\sd}{\mathop{\Delta}}

% Entropy

\newcommand{\mH}{\mathbf{H_{\infty}}}
\newcommand{\avgmH}{\mathbf{\widetilde{H}_{\infty}}}

% Cryptographic elements

\newcommand{\pk}{{pk}}
\newcommand{\evk}{{evk}}
\newcommand{\pp}{{pp}}
\newcommand{\sk}{{sk}}
\newcommand{\vk}{{vk}}
\newcommand{\td}{{td}}

\newcommand{\msk}{{msk}}
\newcommand{\id}{{id}}

\newcommand{\crs}{\text{\sf crs}}

\newcommand{\params}{{params}}
\newcommand{\state}{{setupstate}}

\newcommand{\query}{{query}}
\newcommand{\qstate}{{qstate}}
\newcommand{\resp}{{resp}}

% Algorithms

\newcommand{\keygen}{\mathsf{Keygen}}
\newcommand{\gen}{\mathsf{Gen}}
\newcommand{\eval}{\mathsf{Eval}}
\newcommand{\setup}{\mathsf{Setup}}
\newcommand{\extract}{\mathsf{Extract}}
\newcommand{\enc}{\mathsf{Enc}}
\newcommand{\dec}{\mathsf{Dec}}

\newcommand{\symname}{\mathsf{SYM}}
\newcommand{\symkeygen}{\mathsf{SYM.Keygen}}
\newcommand{\symgen}{\symkeygen}
\newcommand{\symenc}{\mathsf{SYM.Enc}}
\newcommand{\symdec}{\mathsf{SYM.Dec}}

\newcommand{\shname}{\mathsf{SH}}
\newcommand{\shkeygen}{\mathsf{SH.Keygen}}
\newcommand{\shgen}{\shkeygen}
\newcommand{\shenc}{\mathsf{SH.Enc}}
\newcommand{\shdec}{\mathsf{SH.Dec}}
\newcommand{\sheval}{\mathsf{SH.Eval}}

\newcommand{\hename}{\mathsf{HE}}
\newcommand{\hekeygen}{\mathsf{HE.Keygen}}
\newcommand{\hegen}{\hekeygen}
\newcommand{\heenc}{\mathsf{HE.Enc}}
\newcommand{\hedec}{\mathsf{HE.Dec}}
\newcommand{\heeval}{\mathsf{HE.Eval}}

\newcommand{\fhname}{\mathsf{FH}}
\newcommand{\fhkeygen}{\mathsf{FH.Keygen}}
\newcommand{\fhenc}{\mathsf{FH.Enc}}
\newcommand{\fhdec}{\mathsf{FH.Dec}}
\newcommand{\fheval}{\mathsf{FH.Eval}}

\newcommand{\btsname}{\mathsf{BTS}}
\newcommand{\btkeygen}{\mathsf{BTS.Keygen}}
\newcommand{\btgen}{\btkeygen}
\newcommand{\btenc}{\mathsf{BTS.Enc}}
\newcommand{\btdec}{\mathsf{BTS.Dec}}
\newcommand{\bteval}{\mathsf{BTS.Eval}}

\newcommand{\fhename}{\mathsf{FHE}}
\newcommand{\fhekeygen}{\mathsf{FHE.Keygen}}
\newcommand{\fhegen}{\fhekeygen}
\newcommand{\fheenc}{\mathsf{FHE.Enc}}
\newcommand{\fhedec}{\mathsf{FHE.Dec}}
\newcommand{\fheeval}{\mathsf{FHE.Eval}}

\newcommand{\kdmkeygen}{\mathsf{KDM.Keygen}}
\newcommand{\kdmenc}{\mathsf{KDM.Enc}}
\newcommand{\kdmdec}{\mathsf{KDM.Dec}}

\newcommand{\vssgen}{\mathsf{VSS.Gen}}

\newcommand{\pirname}{\mathsf{PIR}}
\newcommand{\pirsetup}{\mathsf{PIR.Setup}}
\newcommand{\pirquery}{\mathsf{PIR.Query}}
\newcommand{\piranswer}{\mathsf{PIR.Response}}
\newcommand{\pirresp}{\piranswer}
\newcommand{\pirdec}{\mathsf{PIR.Decode}}

% Document specific definitions

\newcommand{\idl}[1]{\left\langle{#1}\right\rangle}
\newcommand{\rlwe}{\mathsf{RLWE}}
\newcommand{\plwe}{\mathsf{PLWE}}
\newcommand{\drlwe}{\text{\rm G-RLWE}}
\newcommand{\sdrlwe}{\text{\rm RLWE}}
\newcommand{\vssm}{\text{\rm SVSS}}

\newcommand{\ekdm}{{\cE_{\kdm}}}
\newcommand{\usr}{{\nu}}

\newcommand{\add}{\mathsf{add}}
\newcommand{\mlt}{\mathsf{mult}}

\newcommand{\hc}{\hat{c}}

\newcommand{\otild}{{\widetilde{O}}}
\newcommand{\omtild}{{\widetilde{\Omega}}}

\newcommand{\linf}{{\ell_{\infty}}}

\newcommand{\gf}{{\text{GF}}}

\newcommand{\zset}[1]{\{0, \ldots, {#1}\}}

\newcommand{\fig}[4]{
        \begin{figure}
        \setlength{\epsfysize}{#2}
        \vspace{3mm}
        \centerline{\epsfbox{#4}}
        \caption{#3} \label{#1}
        \end{figure}
        }

\newcommand{\ord}{{\rm ord}}

\providecommand{\norm}[1]{\lVert #1 \rVert}
\newcommand{\embed}{{\rm Embed}}
\newcommand{\qembed}{\mbox{$q$-Embed}}
\newcommand{\lp}{{\rm LP}}